\pdfoutput=1

\documentclass[11pt]{article}

\usepackage[preprint]{acl}

\usepackage{times}
\usepackage{latexsym}

\usepackage[T1]{fontenc}

\usepackage[utf8]{inputenc}

\usepackage{microtype}

\usepackage{inconsolata}

\usepackage{graphicx}

\usepackage{amsmath}
\usepackage{amsfonts}
\usepackage{graphicx}
\usepackage{tcolorbox} 
\usepackage{booktabs} 
\usepackage{ulem} 
\usepackage{caption} 
\usepackage{graphicx} 
\usepackage{algorithm}
\usepackage{algorithmic}
\usepackage{colortbl}
\usepackage{multirow}
\usepackage{bigstrut}
\usepackage{hyperref}

\usepackage{fontawesome}        
\usepackage{xcolor}             
\usepackage{graphicx}           

\usepackage{enumitem}

\title{Table-Critic: A Multi-Agent Framework for \\Collaborative Criticism and Refinement in Table Reasoning}

\author{
 \textbf{Peiying Yu\textsuperscript{1}},
 \textbf{Guoxin Chen\textsuperscript{2,\textsection}},
 \textbf{Jingjing Wang\textsuperscript{1,\textsection}}
\\
 \textsuperscript{1}Natural Language Processing Lab, Soochow University \\
 \textsuperscript{2}Institute of Computing Technology, Chinese Academy of Sciences
\\
 \texttt{\{pyyu@stu,djingwang@\}suda.edu.cn}\quad
 \texttt{chenguoxin22@mails.ucas.ac.cn}
}

\begin{document}
\maketitle
\renewcommand{\thefootnote}{\textsection}
\footnotetext{Corresponding authors}
\renewcommand{\thefootnote}{\arabic{footnote}}
\begin{abstract}
Despite the remarkable capabilities of large language models (LLMs) in various reasoning tasks, they still struggle with table reasoning tasks, particularly in maintaining consistency throughout multi-step reasoning processes.
While existing approaches have explored various decomposition strategies, they often lack effective mechanisms to identify and correct errors in intermediate reasoning steps, leading to cascading error propagation.
To address these issues, we propose Table-Critic, a novel multi-agent framework that facilitates  collaborative criticism and iterative refinement of the reasoning process until convergence to correct solutions. Our framework consists of four specialized agents: a Judge for error identification, a Critic for comprehensive critiques, a Refiner for process improvement, and a Curator for pattern distillation.
To effectively deal with diverse and unpredictable error types, we introduce a self-evolving template tree that systematically accumulates critique knowledge through experience-driven learning and guides future reflections.
Extensive experiments have demonstrated that Table-Critic achieves substantial improvements over existing methods, achieving superior accuracy and error correction rates while maintaining computational efficiency and lower solution degradation rate.
The code is available at \url{https://github.com/Peiying-Yu/Table-Critic}.

\end{abstract}

\begin{figure*}[t]
  \centering
  \includegraphics[width=\textwidth]{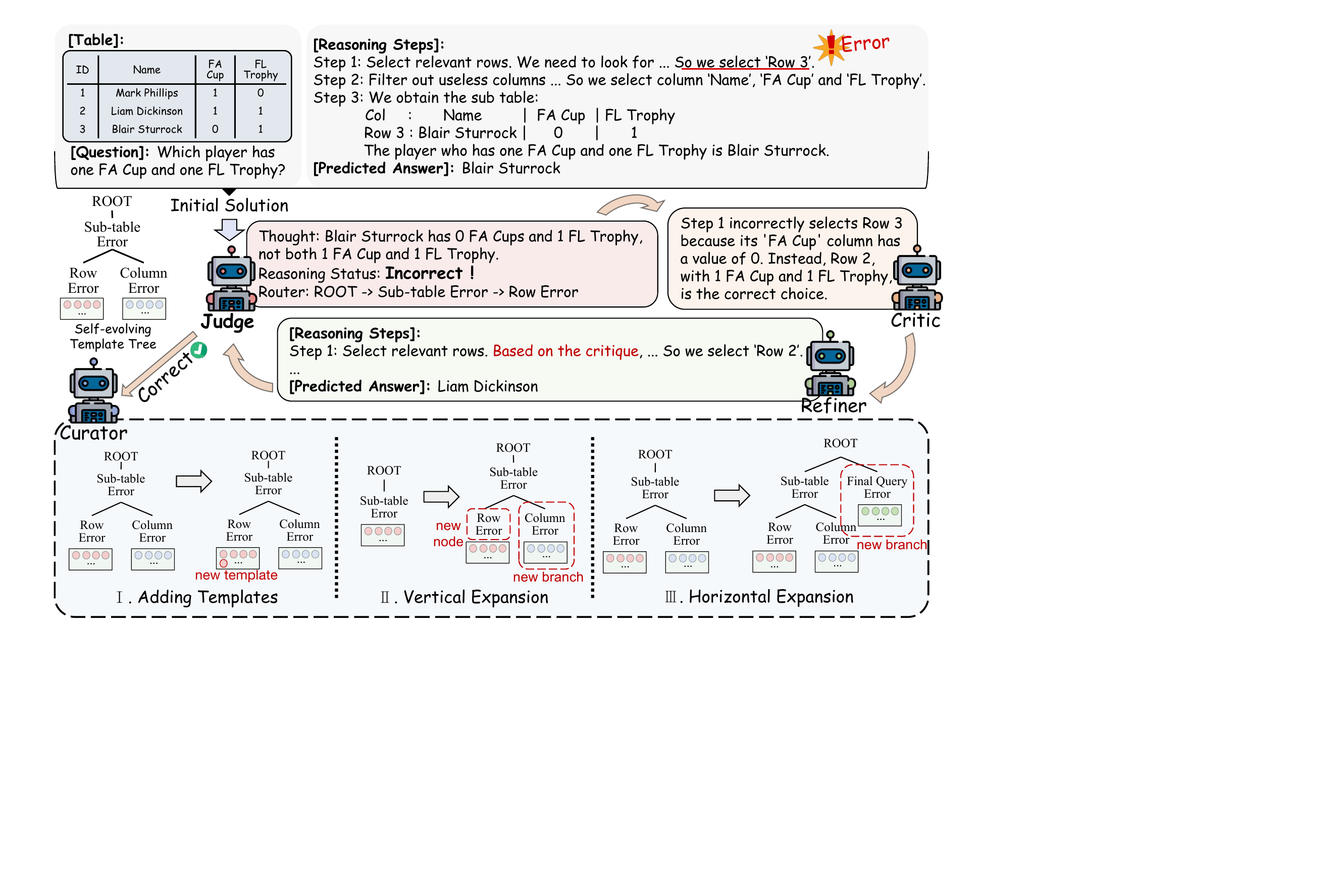}
  \caption{An illustration of Table-Critic, a multi-agent framework for table reasoning tasks, where the Judge identifies errors, the Critic provides detailed critique, the Refiner corrects the reasoning process, and the Curator updates a self-evolving template tree to accumulate critique knowledge and improve future performance.}
  \label{fig:framework}
\end{figure*}

\section{Introduction}
Despite significant advances in various reasoning tasks~\citep{plaat2024reasoning,yu2024natural,alphamath,chen-etal-2024-step,guo2025deepseek,xu2025rap}, large language models (LLMs)~\citep{yang2024qwen2,llama3,claude,gemma,gpt_4o} face substantial challenges in handling semi-structured data, such as table reasoning tasks, as they require both understanding of tabular structures and precise localization of relevant entries in redundant and noisy information~\citep{zhao2024tapera,chen2024tablerag,zhang2025survey}.


Existing approaches address these challenges through various decomposition strategies. 
For example, Binder~\citep{cheng2022binding} decomposes complex questions into executable sub-programs (i.e., SQL or Python), while approaches such as Dater~\citep{ye2023large} and Chain-of-Table~\citep{wang2024chain} focus on dynamic table decomposition for context-aware reasoning.
Although these decomposition-based methods have demonstrated promising performance, they suffer from a critical limitation: the lack of effective mechanisms to criticize and refine the intermediate reasoning steps.
This deficiency inevitably leads to error propagation throughout the reasoning process, significantly affecting the accuracy of final predictions.


However, recent studies~\citep{MadaanTGHGW0DPY23,abs-2412-19513} have revealed that while LLMs possess self-reflection capabilities to some extent, their self-reflection often lacks reliability and consistency.
Simply forcing LLMs to engage in self-reflection may introduce additional biases, especially in table reasoning tasks, wherein models tend to either rationalize their previous erroneous reasoning or over-criticize correct steps, rather than identifying genuine errors~\citep{critic-cot,chen2025learning}.

To address these issues, we propose Table-Critic, a multi-agent framework that introduces specialized agents to collaboratively criticize and refine the reasoning process in a step-by-step manner.
Specifically, our Table-Critic simulates human-like reflective behaviors through four targeted agents: a Judge that identifies potential errors, a Critic that provides detailed suggestions, a Refiner that refines the entire reasoning process, and a Curator that distills critique patterns to guide future reflection.
The collaborative strategy among multiple agents is motivated by our two insights:
\textbf{(1) LLMs demonstrate proficiency in identifying and refining the first erroneous steps, yet tend to make other mistakes in subsequent steps, particularly when dealing with complex problems.}
This observation motivates our multi-turn design where different agents continuously monitor and refine the reasoning process until the Judge verifies its correctness.
\textbf{(2) The diversity and unpredictability of error types in the reasoning process make it challenging for LLMs to effectively identify them based solely on their inherent knowledge.}
This insight motivates the development of a dynamic template repository that categorizes and stores critique templates by error types, allowing our multi-agent system to systematically accumulate critique knowledge.
Specifically, the Curator maintains a self-evolving template tree by expanding branches or adding templates after the entire reflection, while the Judge routes through the tree based on the identified reasoning errors to locate appropriate templates for assisting the Critic in generating high-quality critique, thereby facilitating subsequent refinement.
Through this self-evolving template tree mechanism, our system continuously accumulates and distills critique patterns from previous experiences, enabling more effective error identification beyond LLMs' inherent capabilities.
This experience-driven approach ensures continuous improvement in the quality and consistency.


\textbf{Our contributions} are summarized as follows:
\begin{itemize}[topsep=1pt, partopsep=1pt, leftmargin=12pt, itemsep=-1pt]
    \item We introduce Table-Critic, a novel multi-agent framework where specialized agents collaboratively criticize and refine the reasoning process for complex table reasoning tasks.
    \item We design a multi-turn refinement mechanism where different agents continuously monitor and improve the reasoning process, effectively mitigating error propagation in multi-step reasoning.
    \item We introduce a self-evolving template tree that systematically accumulates and organizes critique knowledge, enabling our system to effectively handle emerging error types through experience-driven learning.
    \item Extensive experiments demonstrate that Table-Critic significantly outperforms existing methods and exhibits substantial advantages over majority voting under comparable or even superior computational costs.
\end{itemize}

\section{Related Work}
\textbf{Table Reasoning.}
Table reasoning, which requires joint understanding of semi-structured tables and questions, has evolved through several paradigms. Early approaches focused on developing specialized models through fine-tuning~\citep{yin2020tabert,liu2021tapex,gu2022pasta}, while recent work has shifted towards leveraging large language models (LLMs) in few-shot learning~\citep{chen2024tablerag,zhao2024tapera}. To handle complex reasoning tasks, decomposition-based methods have emerged as a promising direction. These methods break down complex tasks into manageable steps, either through program execution~\citep{cheng2022binding} or context-aware table partitioning~\citep{ye2023large,wang2024chain}. However, a critical limitation of existing approaches is their inability to effectively critique and refine intermediate reasoning steps, leading to error propagation.
In contrast, our Table-Critic framework addresses this limitation by introducing systematic critique and refinement mechanisms throughout the reasoning process.


\textbf{Self-Reflection.}
Recent studies have revealed that while LLMs possess inherent self-reflection capabilities, they often suffer from reliability and consistency issues~\citep{MadaanTGHGW0DPY23,abs-2412-19513}. Simply enforcing self-reflection can be counterproductive, as models tend to either rationalize their errors or excessively critique correct reasoning steps~\citep{critic-cot,chen2025learning}. To address these limitations, our Table-Critic introduces a structured approach through: (1) a multi-agent framework where specialized agents collaborate to provide targeted critiques, and (2) a self-evolving template tree that systematically accumulates and organizes critique knowledge. This design effectively overcomes the inherent limitations of LLMs' reflection capabilities while maintaining reliable and consistent error identification.

\section{Table-Critic}
\subsection{Overview}
To effectively implement the human-like correction process in multi-step reasoning, we propose a collaborative multi-agent framework, Table-Critic. As illustrated in Figure~\ref{fig:framework}, this framework decomposes the complex reasoning refinement task into four specialized functions: error detection (Judge), critique generation (Critic), reasoning refinement (Refiner), and experience learning (Curator). These agents work in concert to progressively improve reasoning quality while accumulating valuable correction experiences.
Specifically, given a table $\mathbb{T}$ and a question $q$, these agents iteratively refine the initial reasoning chain $\tau = \{s_1, s_2, ..., s_n\}$ until reaching a satisfactory solution.\footnote{We use Chain-of-Table~\citep{wang2024chain} for initial chains, though our framework is applicable to other methods.} 
The refinement process is guided by a self-evolving template tree $\mathcal{T}$ that systematically accumulates critique patterns from past experiences.

\subsection{Multiple Agents}
Inspired by human-like correction behavior, we design four specialized agents---Judge, Critic, Refiner, and Curator---to facilitate criticizing and refining in multi-step reasoning.
We use specific instructions to prompt LLM ($\pi$) to execute the corresponding operations.
Formally, we define each agent as follows:

\textbf{Judge ($\mathcal{A}^j$).}
The Judge agent is responsible for identifying potential errors in the reasoning process.
Given a table $\mathbb{T}$, question $q$, current reasoning chain $\tau$, and the template tree $\mathcal{T}$, it analyzes each reasoning step and determines the specific error type if any exists. Based on the identified error type (if exists), the Judge routes through the template tree $\mathcal{T}$ to locate appropriate templates for guiding the subsequent critic agent.
Formally, the Judge agent operates as:
\begin{equation}
    E, P, R = \pi(\mathbb{T},q,\tau,\mathcal{T},\text{instruction}^{\mathcal{A}^j}),
\end{equation}
where $E$ denotes the error analysis for each reasoning step, $P \in \{\text{Correct}, \text{Incorrect}\}$ indicates the overall reasoning status, and $R$ represents the routing path in the template tree that guides template selection.
Based on the routing path, we sample relevant critique templates $\mathcal{T}_s$ from the template tree $\mathcal{T}$ to guide the Critic agent in generating targeted and high-quality critiques for the identified errors.
Notably, due to the self-evolving nature of our template tree, when the Judge identifies an error type not yet present in the tree, we randomly sample various error types from existing templates to guide the Critic in generating helpful critique.

\textbf{Critic ($\mathcal{A}^c$).}
The Critic agent serves as a crucial component in our framework, responsible for generating detailed and constructive critiques for the identified errors. With the guidance of sampled critique templates $\mathcal{T}_s$, the Critic agent locates the first error step in the reasoning chain $\tau$, analyzes error details, and provides specific suggestions for subsequent refinement. Formally, the Critic agent operates as:
\begin{equation}
    \mathcal{C}, I = \pi(\mathbb{T}, q, \tau, \mathcal{T}_s, \text{instruction}^{\mathcal{A}^c}),
\end{equation}
where $\mathcal{C}$ denotes the generated critique and $I$ indicates the index of the first error step in $\tau$.
The effectiveness of the Critic agent directly impacts the Refiner's ability to correct reasoning errors, which motivates our design of the template tree to enhance critique quality.

\textbf{Refiner ($\mathcal{A}^r$).}
The Refiner agent is tasked with correcting the reasoning chain based on the critique provided by the Critic. Given the critique $\mathcal{C}$, the table $\mathbb{T}$, question $q$, and the partial reasoning chain up to the first error step (i.e., $\tau_p=\{s_1, ..., s_I\}$), the Refiner first rectifies the identified error and then completes the remaining reasoning steps to generate a full refined chain.
Formally, the Refiner agent operates as:
\begin{equation}
    \tau' = \pi(\mathbb{T}, q, \tau_p, \mathcal{C},  \text{instruction}^{\mathcal{A}^r}),
\end{equation}
where $\tau'$ represents the newly generated complete reasoning chain.

\textbf{Curator ($\mathcal{A}^{cu}$).}
The Curator agent serves as an experience-driven learning component that distills valuable critique templates from current refinement processes.
It is activated only after the complete refinement process concludes, specifically when the Judge agent verifies that the final reasoning chain is error-free ($P=\text{Correct}$), as shown in Figure~\ref{fig:framework}.
Through reviewing each refinement iteration and the existing template tree $\mathcal{T}$, the Curator autonomously distills meaningful critique templates from effective refinement experiences. 
These newly distilled templates are then incorporated into $\mathcal{T}$ to enhance future critique generation. 
Formally, the Curator operates as:
\begin{equation}
    \mathcal{T}' = \pi(\mathcal{T}, H, \text{instruction}^{\mathcal{A}^{cu}}),
\end{equation}
where $H$ represents the complete refinement history, and $\mathcal{T}'$ denotes the updated template tree.
The detailed update strategy will be delineated in subsequent sections.

\subsection{Multi-turn Refinement}
As discussed in the Introduction, the multi-turn refinement in Table-Critic is motivated by our observation that LLMs often excel at identifying and correcting the first error in reasoning chains, but may introduce new errors in subsequent steps. To address this challenge, we implement an iterative refinement process where multiple agents collaboratively monitor and improve the reasoning chain until reaching a satisfactory solution.

Specifically, given an initial reasoning chain $\tau$, our framework operates through the following steps in each iteration:
(1) The Judge agent first analyzes the entire reasoning chain to identify potential errors and determine their types. If no errors are detected ($P=\text{Correct}$), the process terminates. Otherwise, the Judge routes through the template tree to locate relevant critique templates.
(2) With the guidance of sampled templates $\mathcal{T}_s$, the Critic agent generates detailed critiques $\mathcal{C}$ focusing on the first identified error at step $I$. This strategy ensures that each refinement iteration addresses errors sequentially, preventing the introduction of cascading errors.
(3) The Refiner agent then generates a new reasoning chain $\tau'$ by incorporating the critique. Importantly, the Refiner only receives the partial chain $\tau_p$ up to the error step $I$, forcing it to reconstruct the remaining steps with the help of critique. This design prevents the Refiner from being biased by previous erroneous chain.
(4) The above process continues iteratively until one of the following conditions is met: the Judge determines the current reasoning chain is correct ($P=\text{Correct}$) or the maximum number of iterations $K$ is reached.

Through this multi-turn design, Table-Critic effectively manages the complexity of multi-step reasoning refinement while maintaining the quality of each correction step. The iterative nature of our approach, combined with specialized agent roles and strategic process control, enables robust and efficient reasoning improvement.

\subsection{Self-evolving Template Tree}
To address the challenge of identifying diverse and unpredictable error types in table reasoning, we introduce a self-evolving template tree that systematically accumulates and organizes critique knowledge. This dynamic structure enables our system to effectively handle both common and emerging error patterns through experience-driven learning.

\textbf{Tree Structure.}
The template tree $\mathcal{T}$ represents a hierarchical structure that captures the relationships among different error types.
As shown in Figure~\ref{fig:framework}, each node in the tree represents a specific type of error, where: (1) Internal nodes represent broader error categories (e.g., Sub-table Error) that can be further subdivided into more specific error types.
(2) Leaf nodes represent specific error types (e.g., Row Error, Column Error) and maintain a repository of critique templates associated with that particular error type.

\textbf{Self-evolving Mechanism.} 
The template tree evolves dynamically through the Curator agent, which manages two primary operations: adding templates to existing leaf nodes and expanding tree branches. As illustrated in Figure~\ref{fig:framework}, the evolution process includes:

(1) \textit{Template Enhancement}. When new effective critique patterns are identified, the Curator adds them to the corresponding leaf node's template repository. This operation enriches existing error type categories without changing the tree structure. For instance, when a new effective template for Row Error is discovered, it is directly added to the corresponding template repository.

(2) \textit{Branch Expansion}. The Curator expands the tree structure in two ways when new error types are identified:

\begin{itemize}[topsep=1pt, partopsep=1pt, leftmargin=12pt, itemsep=-1pt]
    \item \textit{Vertical Expansion}: When a new error type is discovered that requires more fine-grained categorization, the Curator performs a vertical split. This operation transforms an existing leaf node into an internal node with two new child nodes. Specifically, the Curator first categorizes the existing templates in the leaf node with an appropriate name (e.g., Row Error), creating one new leaf node. Then, it creates another leaf node with a different name (e.g., Column Error) to accommodate the newly discovered error type and its corresponding templates. This process ensures that each leaf node maintains a cohesive collection of templates for a specific error type.
    
    \item \textit{Horizontal Expansion}: When a completely new error type is identified that parallels existing categories, the Curator adds a new branch at the same level. This operation preserves the existing structure while accommodating new error types. As illustrated in the Figure~\ref{fig:framework} (bottom), the addition of the Final Query Error branch represents a horizontal expansion that complements the existing Sub-table Error category.
\end{itemize}

Through these evolution mechanisms, our template tree maintains a dynamic balance between preserving accumulated knowledge and incorporating new error patterns. The vertical expansion enables more precise error categorization, while horizontal expansion ensures comprehensive coverage of diverse error types. This adaptive structure allows the system to continuously improve its critique capabilities while maintaining organized and efficient template management.
The detailed pipeline of our Table-Critic is presented in Appendix~\ref{app:algorithm}.

\definecolor{lightpurple}{RGB}{216,236,228}

\begin{table*}[htbp]
    \centering
    \resizebox{\textwidth}{!}{
        \begin{tabular}{lcccccccc}
            \toprule 
            \multirow{2}{*}{\textbf{Method}} & \multicolumn{2}{c}{\textbf{Qwen2.5-72B}} & \multicolumn{2}{c}{\textbf{LLaMA3.3-70B}} & \multicolumn{2}{c}{\textbf{GPT-4o-mini}} & \multicolumn{2}{c}{\textbf{Average}} \\
            \cmidrule(lr){2-3} \cmidrule(lr){4-5} \cmidrule(lr){6-7} \cmidrule(lr){8-9}
            & WikiTQ & TabFact & WikiTQ & TabFact & WikiTQ & TabFact & WikiTQ & TabFact \\
            \midrule 
            End-to-End QA & 56.6 & 85.1 & 51.1 & 81.0 & 52.6 & 73.5 & 53.4 & 79.9 \\
            Few-Shot QA & 61.7 & 85.0 & 62.0 & 80.7 & 57.6 & 75.1 & 60.4 & 80.3 \\
            Binder \citep{cheng2022binding} & 57.0 & 82.2 & 52.2 & 80.5 & 54.8 & 83.3 & 54.7 & 82.0 \\
            Dater \citep{ye2023large} & 63.8 & \uline{90.0} & 59.5 & 87.6 & 65.8 & 83.6 & 63.0 & 87.1 \\
            Chain-of-Table \citep{wang2024chain} & 68.3 & 89.7 & 62.1 & \uline{89.9} & \uline{67.5} & \uline{88.9} & 66.0 & \uline{89.5} \\
            Critic-CoT \citep{critic-cot} & \uline{69.0} & 89.8 & \uline{66.8} & 88.0 & 66.3 & 86.9 & \uline{67.4} & 88.2 \\
            \midrule 
            \rowcolor{lightpurple} 
            Table-Critic (ours) & \textbf{77.2} & \textbf{92.6} & \textbf{70.1} & \textbf{91.5} & \textbf{73.9} & \textbf{91.1} & \textbf{73.7} & \textbf{91.7} \\
             \rowcolor{lightpurple}
             & \textcolor{red}{\(\uparrow\)8.2} & \textcolor{red}{\(\uparrow\)2.6} & \textcolor{red}{\(\uparrow\)3.3} & \textcolor{red}{\(\uparrow\)1.6} & \textcolor{red}{\(\uparrow\)6.4} & \textcolor{red}{\(\uparrow\)2.2} 
             & \textcolor{red}{\(\uparrow\)6.3} & \textcolor{red}{\(\uparrow\)2.2} \\
            \bottomrule 
        \end{tabular}
}
\caption{Table reasoning results on WikiTQ and TabFact with Qwen2.5-72B, LLaMA3.3-70B, and GPT-4o-mini. Bold denotes the best performance and underline denotes the second-best performance.}
\vspace{-0.8em}
\label{mainresult}
\end{table*}

\section{Experiments}
\label{experiments}

\subsection{Experimental Setup}
\textbf{Datasets.} We evaluate our approach on two standard benchmarks:
(1) WikiTableQuestions (WikiTQ)~\citep{pasupat2015compositional}: A table reasoning benchmark with 4,344 test samples from 421 tables.
(2) TabFact~\citep{chen2019tabfact}: A fact verification benchmark in table reasoning with 2,024 test samples from 298 tables.

\textbf{Baselines.} We conduct comprehensive experiments comparing Table-Critic against three categories of baselines:
\textbf{(1) Standard Reasoning.} \text{End-to-End QA} directly generates answers using table and question as input. \text{Few-Shot QA} extends this by incorporating exemplar (Table, Question, Answer) triplets from the training set.
\textbf{(2) Decomposition-Based Reasoning.} Binder~\citep{cheng2022binding} decomposes questions into executable SQL/Python sub-programs. Dater~\citep{ye2023large} employs parsing-execution-filling strategy with sub-table decomposition. Chain-of-Table~\citep{wang2024chain} generates intermediate tables through dynamic operations.
\textbf{(3) Critic-Based Reasoning.} Critic-CoT~\citep{critic-cot} implements self-reflection for error identification.

\textbf{Implementation Details.}
To ensure comprehensive evaluation, we conduct experiments across three LLMs: Qwen2.5-72B-Instruct~\citep{yang2024qwen2}, LLaMA3.3-70B-Instruct~\citep{llama3}, and GPT-4o-mini~\citep{gpt_4o}.
For all baseline methods, we follow their original settings to ensure optimal performance.
For fair comparison, both Critic-CoT~\citep{critic-cot} and our Table-Critic framework are implemented upon Chain-of-Table~\citep{wang2024chain}.
For our Table-Critic, the template tree is initialized with only 2 templates that demonstrate basic critique patterns. From this minimal starting point, the tree evolves autonomously through our self-evolving mechanism, continuously learning and incorporating new critique patterns.
For all experiments, we set the maximum refinement iterations K to 5 and use temperature 0.0 for greedy decoding. The detailed prompts and instructions for each agent in our framework are provided in Appendix~\ref{app:prompts}.

\subsection{Main Results}
We report the performance on different table reasoning benchmarks across different LLMs in Table~\ref{mainresult}.
Our comprehensive evaluation reveals several key findings:
\textbf{First,} Table-Critic consistently outperforms all baseline methods across both datasets and all three LLMs. On average, our method achieves 73.7\% accuracy on WikiTQ and 91.7\% on TabFact, representing significant improvements of 6.3\% and 2.2\% respectively over the strongest baselines.
\textbf{Second,} the improvements are robust across different model architectures. With Qwen2.5-72B-Instruct, we achieve the highest absolute performance (77.2\% on WikiTQ, 92.6\% on TabFact), showing substantial gains of 8.2\% and 2.6\% respectively. Similar patterns are observed with LLaMA3.3-70B-Instruct and GPT-4o-mini, demonstrating the framework's generalizability across different foundation models.
\textbf{Third,} the performance variations between WikiTQ and TabFact provide insights into our method's strengths. Table-Critic shows larger improvements on WikiTQ (average +6.3\%) compared to TabFact (average +2.2\%), indicating its particular effectiveness in handling complex, multi-step reasoning tasks. This aligns with our framework design, as WikiTQ's compositional questions benefit more from our multi-turn refinement and self-evolving template tree mechanism than TabFact's binary verification tasks. Nevertheless, the consistent improvements on TabFact demonstrate our method's capability even in simpler scenarios.
\textbf{Finally,} comparing against different baseline categories reveals the advancement of our approach. While recent methods like Chain-of-Table~\citep{wang2024chain} and Critic-CoT~\citep{critic-cot} have made notable progress through decomposition and criticism mechanisms, Table-Critic achieves substantially larger improvements over these strong baselines. This suggests that our multi-agent framework, combining multi-turn refinement with self-evolving template tree, provides a more effective solution for complex table reasoning tasks.

\subsection{Analysis of Critic Effectiveness}

\begin{table*}[htbp]
    \centering
    \begin{tabular}{lccccccccc}
        \toprule
        \multirow{2}{*}{\textbf{Dataset}} & \multicolumn{1}{c}{\textbf{Chain-of-Table}} & \multicolumn{4}{c}{\textbf{Critic-CoT}} & \multicolumn{4}{c}{\textbf{Table-Critic}} \\
        \cmidrule(lr){2-2} \cmidrule(lr){3-6} \cmidrule(lr){7-10}
        & Acc & Acc & $\Delta^{\text{i}\to\text{c}}$ & $\Delta^{\text{c}\to\text{i}}$ & $\Delta$ & Acc & $\Delta^{\text{i}\to\text{c}}$ & $\Delta^{\text{c}\to\text{i}}$ & $\Delta$ \\
        \midrule
        WikiTQ & 68.3 & 69.0 & +5.6 & -4.9 & \textcolor{red}{+0.7} & \cellcolor{lightpurple}77.2 & \cellcolor{lightpurple}+9.6 & \cellcolor{lightpurple}-0.7 & \cellcolor{lightpurple}\textcolor{red}{+8.9} \\
        TabFact & 89.7 & 89.8 & +2.9 & -2.8 & \textcolor{red}{+0.1} & \cellcolor{lightpurple}92.6 & \cellcolor{lightpurple}+3.4 & \cellcolor{lightpurple}-0.5 & \cellcolor{lightpurple}\textcolor{red}{+2.9} \\
        \bottomrule
    \end{tabular}
    \caption{Critic performance comparison of different critic methods. $\Delta^{\text{i}\to\text{c}}$, $\Delta^{\text{c}\to\text{i}}$, and $\Delta$ measure the error correction rate, solution degradation rate, and net performance gain relative to Chain-of-Table, respectively.}
    \vspace{-1em}
    \label{tab:critic}
\end{table*}


As shown in Table~\ref{tab:critic}, we conduct a detailed analysis of different critic mechanisms by comparing Table-Critic with Chain-of-Table~\citep{wang2024chain} and Critic-CoT~\citep{critic-cot}. Our analysis focuses on four key metrics: 
\textbf{(1) Overall Accuracy (Acc)}: The percentage of correctly solved questions;
\textbf{(2) Error Correction Rate ($\Delta^{\text{i}\to\text{c}}$):} The percentage of questions incorrectly solved by Chain-of-Table but corrected by different Critic methods;
\textbf{(3) Solution Degradation Rate ($\Delta^{\text{c}\to\text{i}}$):} The percentage of questions correctly solved by Chain-of-Table but degraded by different Critic methods;
\textbf{(4) Net Performance Gain ($\Delta$):} The overall improvement relative to Chain-of-Table, calculated as $\Delta = \Delta^{\text{i}\to\text{c}} + \Delta^{\text{c}\to\text{i}}$.

\textbf{Error Correction vs. Solution Degradation.}
Table-Critic demonstrates superior error correction capabilities while minimizing solution degradation. On WikiTQ, it successfully corrects 9.6\% of Chain-of-Table's errors while only degrading 0.7\% of correct solutions, resulting in a substantial net performance gain (\text{+8.9\%}). In contrast, Critic-CoT shows a less effective pattern, with a 5.6\% correction rate offset by a high degradation rate (-4.9\%), yielding only a marginal improvement (\text{+0.7\%}).

\textbf{Task-Specific Performance.} The effectiveness of critique mechanisms varies across different tasks. On WikiTQ, which involves complex multi-step reasoning, Table-Critic achieves a higher error correction rate (+9.6\% vs +5.6\%) and maintains a observably lower degradation (-0.7\% vs -4.9\%) compared to Critic-CoT. For TabFact's simpler verification tasks, while the improvements are more modest, Table-Critic still maintains better stability with lower degradation rates (-0.5\% vs -2.8\%).

\textbf{Critic Stability.} A key advantage of Table-Critic is its stability in maintaining correct solutions. The consistently low degradation rates (-0.7\% for WikiTQ and -0.5\% for TabFact) suggest that our self-evolving template tree effectively preserves valid reasoning patterns while identifying and correcting errors. This contrasts with Critic-CoT's higher degradation rates (-4.9\% for WikiTQ and -2.8\% for TabFact), indicating potential instability in its critique process.

\begin{figure}[t]
    \centering
    \begin{minipage}[t]{0.47\textwidth}
        \centering
        \includegraphics[width=\textwidth]{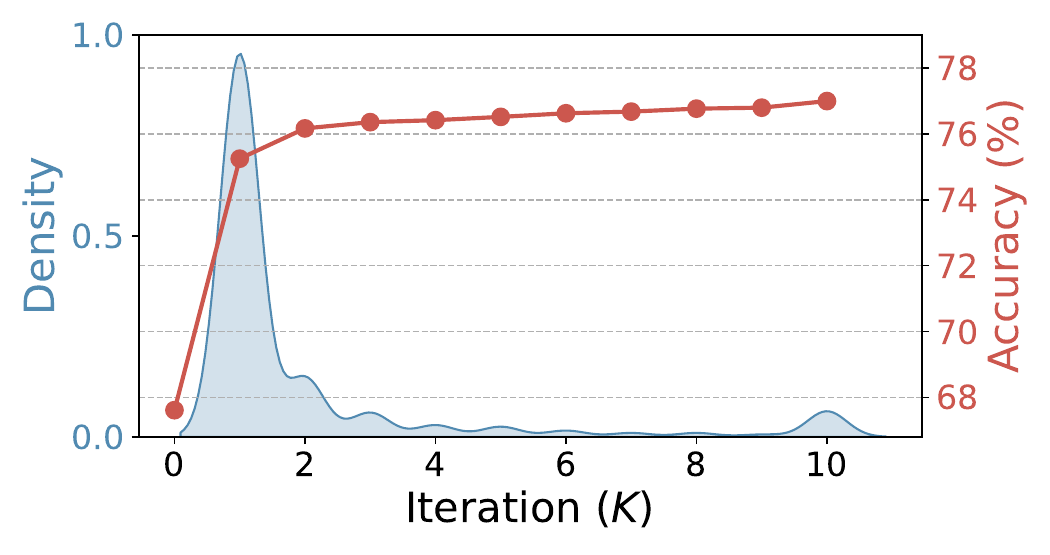}
        \text{(a) WikiTQ}
        \vspace{1pt}
    \end{minipage}

    \begin{minipage}[t]{0.49\textwidth}
        \centering
        \includegraphics[width=\textwidth]{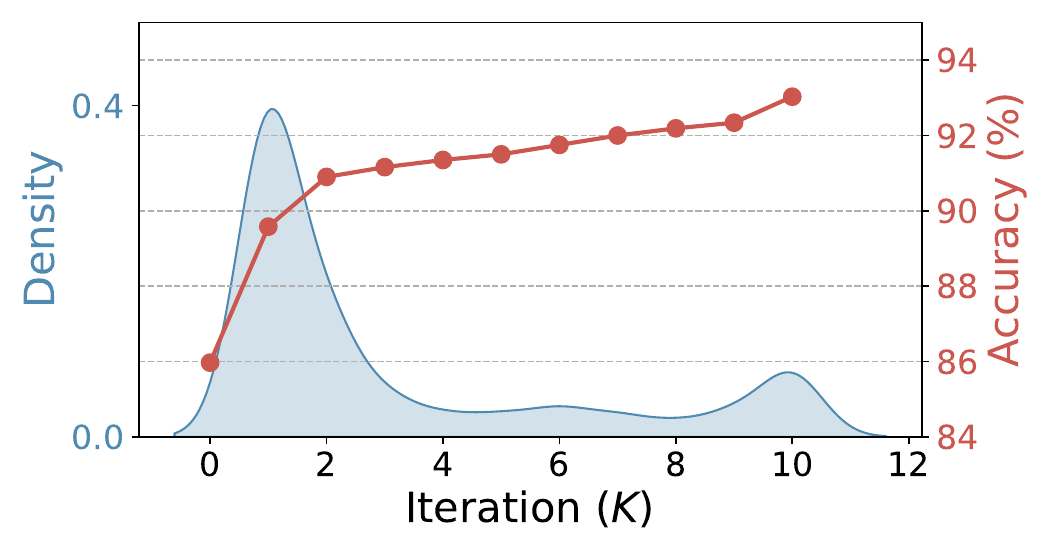}
        \vspace{-3pt}
        \text{(b) TabFact}
    \end{minipage}
    \caption{Analysis of Model Convergence and Iteration Requirements on WikiTQ and TabFact Datasets.}
    \vspace{-0.8em}
    \label{fig:accuracy-density}
\end{figure}

\subsection{Analysis of Multi-Turn Mechanism}
To understand the effectiveness of our multi-turn refinement mechanism, we analyze how model performance evolves with the number of iterations $K$ and the distribution of required iteration counts (set maximal $K=10$), as shown in Figure~\ref{fig:accuracy-density}.

\textbf{Performance Evolution.} On both datasets, we observe a consistent pattern of rapid initial improvement followed by gradual convergence. For WikiTQ, the accuracy increases sharply from 67.6\% to 76.5\% within the first three iterations and stabilizes around 77\% after six iterations. Similarly, on TabFact, the performance improves significantly in early iterations and plateaus at approximately 92\% after five iterations. This pattern suggests that our multi-turn mechanism effectively refines solutions through iterative improvements.

\textbf{Iteration Distribution.} The density plots reveal interesting insights about the complexity of different tasks. On WikiTQ, we observe a broader distribution with multiple peaks, indicating that questions require varying numbers of iterations for resolution. The main peak occurs at 1-2 iterations, with smaller peaks extending up to 10 iterations, reflecting the diverse complexity of multi-step reasoning questions.
TabFact also shows a concentrated distribution with two distinct peaks: a primary peak at 1-2 iterations and a secondary peak around 10 iterations. This bimodal pattern suggests that TabFact tend to fall into two categories: (1) straightforward cases that can be verified quickly within 1-2 iterations, and (2) complex cases that require extensive refinement to reach a conclusive verification. This distribution aligns with the inherent nature of fact verification tasks, where statements are either relatively simple to verify or require careful step-by-step examination.

\textbf{Convergence and Stability Analysis.} The results suggest that while our method allows for up to 10 iterations, most improvements are achieved within the first 5 iterations. This efficient convergence, combined with our early termination mechanism, helps maintain computational efficiency while ensuring thorough reasoning. Notably, as evidenced in Table~\ref{tab:critic}, Table-Critic maintains stable performance across iterations without the degradation typically seen in iterative approaches, demonstrating the effectiveness of our Critic agent and self-evolving template tree mechanism.

\subsection{Analysis of Computational Cost}
\begin{figure}[t]
    \centering
    \begin{minipage}{0.25\textwidth}
        \centering
        \includegraphics[width=\textwidth]{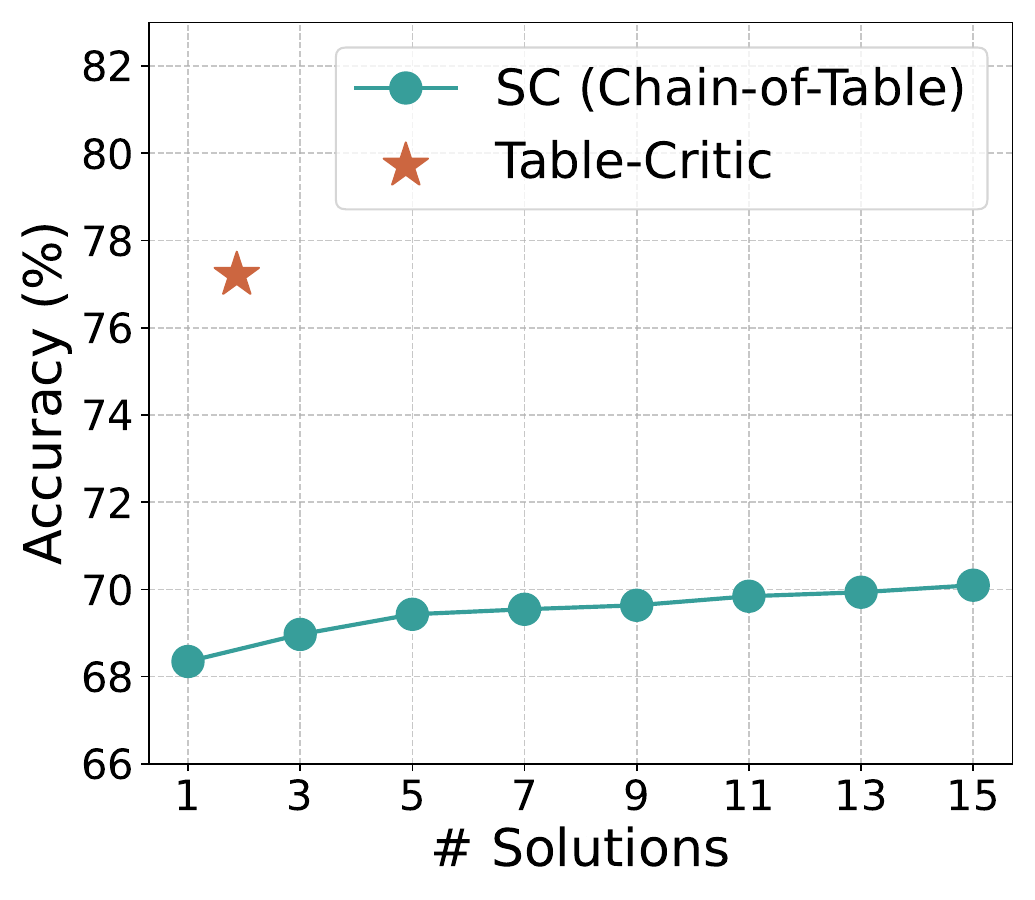}
        \text{(a) WikiTQ}
    \end{minipage}%
    \begin{minipage}{0.25\textwidth}
        \centering
        \includegraphics[width=\textwidth]{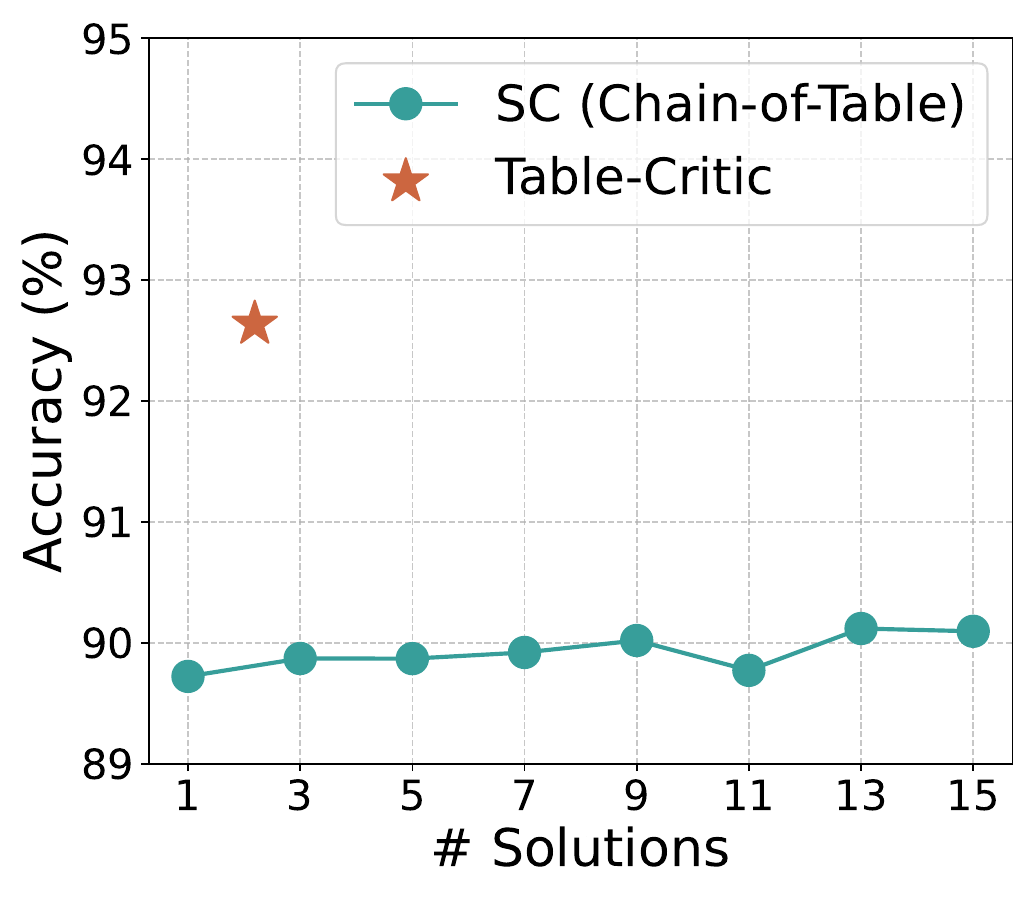}
        \text{(b) TabFact}
    \end{minipage}
    \vspace{-0.2em}
    \caption{Computational cost and Effectiveness Comparison between SC (Self-Consistency Based on Chain-of-Table) and our Table-Critic.}
    \vspace{-1em}
    \label{fig:computation}
\end{figure}

To ensure a fair comparison with Chain-of-Table~\citep{wang2024chain} in terms of computational cost, we conduct an analysis of the cost-effectiveness trade-off, as shown in Figure~\ref{fig:computation}. Since Table-Critic builds upon Chain-of-Table by incorporating additional critique mechanisms, we align the computational costs by allowing Chain-of-Table to generate multiple solutions (majority voting) through Self-consistency~\citep{wang2023selfconsistency} (with temperature 0.8) and compare the performance under equivalent or even superior computational budgets.

\textbf{Efficiency Comparison.} Our method requires approximately 1.8-2.2× computational cost compared to the basic Chain-of-Table.
The complete derivation process of computational cost is provided in Appendix~\ref{app:computation}. 
However, as illustrated in Figure~\ref{fig:computation}, Table-Critic achieves substantially higher accuracy (77.2\% on WikiTQ and 92.6\% on TabFact) compared to Chain-of-Table's performance even with 15 solution attempts. Notably, Chain-of-Table shows only marginal improvements as the number of solutions increases, reaching 70.0\% on WikiTQ and 90.1\% on TabFact with 15 solutions.

\textbf{Cost-Effectiveness Analysis.} The results demonstrate that simply increasing the number of solution attempts in Chain-of-Table fails to achieve comparable performance to Table-Critic, despite consuming similar or even greater computational resources. This suggests that our multi-agent refinement mechanism provides a more effective approach to improving reasoning accuracy than traditional majority voting strategies. The superior performance of Table-Critic justifies its additional computational overhead by offering substantially better reasoning capabilities.

\subsection{Analysis of Self-evolving Template Tree}
To investigate the effectiveness of our self-evolving mechanism, we conduct an ablation study comparing Table-Critic with and without the dynamic template evolution capability, as shown in Table~\ref{tab:ablation}. In the static setting (w/o Self-evolving), the template tree remains fixed with its initial two templates, while our full Table-Critic allows the Curator agent to dynamically maintain and evolve the template tree throughout the reasoning process.

\textbf{Performance Impact.} The results demonstrate the clear benefits of the self-evolving mechanism. Without template evolution, performance drops by 1.1\% on WikiTQ (from 77.2\% to 76.1\%) and 1.8\% on TabFact (from 92.6\% to 90.8\%). The more substantial performance gap on TabFact suggests that template evolution is particularly beneficial for fact verification tasks, where diverse verification patterns may be needed.

\textbf{Mechanism Analysis.} These results highlight the importance of dynamic adaptation in our framework. The self-evolving mechanism allows the template tree to expand beyond its initial state, accommodating diverse reasoning patterns encountered during the critique process. This flexibility enables more effective error detection and correction compared to a static template approach. The performance gains validate our design choice of incorporating dynamic template evolution, showing that the ability to adapt and expand the template structure is crucial for robust table reasoning. For reference, we provide visualizations of both the initial template tree and its evolved state in Appendix~\ref{app:tree}, illustrating how the tree structure adapts to accommodate different reasoning patterns.

\begin{table}[t]
\centering
\begin{tabular}{@{}lcc@{}}
\toprule
\textbf{Method} & \textbf{WikiTQ} & \textbf{Tabfact} \\ \midrule
 Table-Critic & 77.2  & 92.6 \\ 
\quad \text{w/o} Self-evolving   & 76.1  & 90.8 \\
                  & \textcolor{blue}{\(\downarrow\)1.1} & \textcolor{blue}{\(\downarrow\)1.8} \\ \bottomrule
\end{tabular}
\vspace{-0.2em}
\caption{The impact of self-evolving mechanism on our template tree.}
\vspace{-0.8em}
\label{tab:ablation}
\end{table}

\section{Conclusion}
In this paper, we propose Table-Critic, a novel multi-agent framework that enhances table reasoning through collaborative criticism and refinement. Our approach introduces four specialized agents working in concert with a self-evolving template tree, effectively addressing the challenges of error identification and correction in complex table reasoning tasks. Extensive experiments demonstrate that our method significantly outperforms existing approaches, achieving substantial improvements across different datasets while maintaining robust performance stability. 





\section*{Limitations}
Our Table-Critic framework has demonstrated strong performance in enhancing table reasoning through multi-agent collaboration and systematic refinement. While our current implementation focuses primarily on textual table reasoning, the proposed multi-agent critique framework is inherently flexible and can potentially be extended to various other scenarios. For instance, the framework could be adapted to handle multimodal reasoning tasks where tables are combined with images, graphs, or other visual elements. We believe the core principles of our approach—collaborative criticism, iterative refinement, and self-evolving template tree—could contribute to broader applications in complex reasoning tasks beyond the current textual domain.

\section*{Acknowledgments}

This work was supported by three NSFC grants, i.e., No.62006166, No.62376178 and No.62076175. This work was also supported by Collaborative Innovation Center of Novel Software Technology and Industrialization, and supported by a Project Funded by the Priority Academic Program Development of Jiangsu Higher Education Institutions (PAPD).

\bibliography{tact}

\appendix
\newpage

\section{Additional Related works}
\textbf{Multi-agent Systems.}
Multi-agent systems have recently demonstrated promising potential in complex reasoning tasks by enabling collaborative problem-solving through specialized agents~\citep{GuoCWCPCW024,pmlr-v235-zhang24au,guan-greene-2024-advancing,chen2025c}. These systems typically leverage the complementary strengths of different agents to achieve more robust and effective solutions than single-agent approaches. While existing work has explored multi-agent frameworks in various domains~\citep{guan-etal-2024-effective, guan2025prepocr}, their application to table reasoning tasks remains largely unexplored.
To our knowledge, our Table-Critic presents the first attempt to introduce a multi-agent framework for table reasoning, where specialized agents collaborate to identify, critique, and refine reasoning steps, offering a novel perspective on addressing the challenges in complex table reasoning tasks.

\section{More Implementation Details}\label{app:algorithm}
In this section, we provide a comprehensive implementation details of our proposed method. For additional insights and more intricate details, we refer the reader to our \textbf{supplementary materials}.

\subsection{Overall Pipeline of Table-Critic}

Table-Critic employs an iterative process to critique and refine the reasoning chain and predicted answer for table reasoning tasks. As described in Algorithm~\ref{alg1}, the process begins with an input table $\mathbb{T}$, a question $q$, an initial reasoning chain $\tau$, and a template tree $\mathcal{T}$. The \texttt{Judge} agent is first invoked to evaluate the correctness of the reasoning chain (Line 2). This evaluation yields the reasoning status $P$, an error analysis $E$, and a routing path $R$ in the template tree.

When the reasoning chain is deemed incorrect ($P = \text{Incorrect}$), Table-Critic proceeds by sampling relevant critique templates $\mathcal{T}_s$ from the template tree using the routing path $R$ (Line 4). These templates are then used by the \texttt{Critic} agent to generate a detailed critique $\mathcal{C}$ and identify the index of the first error step $I$ in the reasoning chain (Line 5). To address the identified errors, the \texttt{Refiner} agent retains the reasoning steps up to step $I$ and refines the chain starting from step $I$, guided by the critique $\mathcal{C}$ (Line 6). The refined reasoning chain $\tau'$ is subsequently re-evaluated by the \texttt{Judge} agent to determine if it is now correct (Line 7).

This refinement loop continues iteratively until the reasoning chain is verified as correct ($P = \text{Correct}$). Once a correct reasoning chain is established, the \texttt{Curator} agent updates the template tree $\mathcal{T}$ by incorporating new critique templates distilled from the refinement history. This update enhances the template tree's ability to support future refinement processes (Line 9).

The final output of Table-Critic is the refined reasoning chain $\tau'$, which represents the accurate and improved solution to the table reasoning task. By systematically identifying and addressing errors in a collaborative multi-step process, Table-Critic ensures robust and precise refinement of reasoning chains and answers.

\begin{algorithm*}
\renewcommand{\algorithmicrequire}{\textbf{Input:}}
\renewcommand{\algorithmicensure}{\textbf{Output:}}
\caption{The overall pipeline of Table-Critic}
\label{alg1}
\begin{algorithmic}[1]
\REQUIRE Table $\mathbb{T}$, question $q$, initial reasoning chain $\tau$, the template tree $\mathcal{T}$.
\ENSURE The refined reasoning chain $\tau'$.
\STATE $H \leftarrow \emptyset$ {\small \hfill $\triangleright$ Initialize refinement history.}
\STATE $P, E, R \leftarrow \texttt{Judge}(\mathbb{T}, q, \tau, \mathcal{T})$ 
\WHILE{$P = \text{Incorrect}$}
    \STATE $\mathcal{T}_s \leftarrow$ Sample Templates using $R$ in the $\mathcal{T}$ 
    \STATE $\mathcal{C}, I \leftarrow \texttt{Critic}(\mathbb{T}, q, \tau, \mathcal{T}_s)$ \hfill {\small $\triangleright$ Generating critique $\mathcal{C}$ and identify the index of first error step $I$.}
    \STATE $\tau_p \leftarrow \tau[:I]$ \hfill {\small $\triangleright$ Retain the partial reasoning steps up to step $I$}
    \STATE $\tau' \leftarrow \texttt{Refiner}(\mathbb{T}, q, \tau_p, \mathcal{C})$ \hfill {\small $\triangleright$ Refine the reasoning chain.}
    \STATE $H\leftarrow H \cup \{\mathbb{T}, q, \tau, \tau', \mathcal{C}\}$ \hfill {\small $\triangleright$ Update history.}
    \STATE $P, E, R \leftarrow \texttt{Judge}(\mathbb{T}, q, \tau', \mathcal{T})$ \hfill {\small $\triangleright$ Re-evaluates the updated reasoning chain}
\ENDWHILE
\STATE $\mathcal{T} \leftarrow \texttt{Curator}(\mathcal{T}, H)$ \hfill {\small $\triangleright$ Update the template tree $\mathcal{T}$ to facilitate future refinement.}
\RETURN Refined reasoning chain $\tau'$ and new template tree $\mathcal{T}$.
\end{algorithmic}
\end{algorithm*}


\subsection{LLM Servers}
Our approach implements agent behaviors through in-context learning, requiring no extensive training procedures. We deploy multiple LLM servers, including Qwen2.5-72B-Instruct and LLaMA3.3-70B-Instruct through the \href{https://docs.sglang.ai/}{\texttt{SGLang}} inference engine, and GPT-4o-mini via its provided API service. While the choice between fine-tuning and in-context learning remains an open question, it is not the primary focus of our work. Following prior studies~\citep{wang2024chain,critic-cot}, we adopt in-context learning as our implementation strategy for its simplicity and effectiveness.

\section{Detailed Computational Cost Analysis}
\label{app:computation}

\begin{table*}[h]
  \centering
  \begin{tabular}{lccccccc}
    \toprule
    Dataset & 
    \multicolumn{3}{c}{Chain-of-Table} & 
    \multicolumn{3}{c}{Table-Critic} & 
    Cost Ratio \\
    \cmidrule(lr){2-4} \cmidrule(lr){5-7} \cmidrule(lr){8-8}
    & Input (M) & Output (M) & Total (M) & Input (M) & Output (M) & Total (M) & (TC/CoT) \\
    \midrule
    WikiTQ & 73.5 & 1.6 & 19.6 & 135.5 & 3.8 & 36.7 & 1.87× \\
    TabFact & 29.3 & 0.6 & 7.8 & 62.1 & 20.4 & 17.1 & 2.19× \\
    \bottomrule
  \end{tabular}
  \caption{Computational Cost Comparison Between Chain-of-Table and Table-Critic (Token Counts in Millions)}
  \label{cost_comparison}
\end{table*}



This appendix evaluates the computational cost of Table-Critic relative to the baseline Chain-of-Table method. The computational cost is analyzed for two datasets, WikiTQ and TabFact, based on the number of input and output tokens required. All token counts are expressed in millions (M), and the cost ratio reflects the relative cost of Table-Critic compared to Chain-of-Table.

\subsection{Computational Cost Definition}
\label{methodology}

The computational cost of a prompt-based method is defined as follows:
\begin{equation}
    N_{\text{total}} = N_{\text{in}} \cdot \left(\frac{P_{\text{in}}}{P_{\text{in}} + P_{\text{out}}}\right) + N_{\text{out}} \cdot \left(\frac{P_{\text{out}}}{P_{\text{in}} + P_{\text{out}}}\right),
\end{equation}
where $N_{\text{in}}$ and $N_{\text{out}}$ represent the number of input and output tokens, and $P_{\text{in}}$ and $P_{\text{out}}$ denote the costs per token for input and output, respectively. Based on the pricing model of Qwen2.5-72B-Instruct, $P_{\text{in}} = 0.004$ CNY per thousand tokens and $P_{\text{out}} = 0.012$ CNY per thousand tokens.

Using the above values, the normalized cost weights are:
\begin{align*}
    \text{Input Weight} &= \frac{P_{\text{in}}}{P_{\text{in}} + P_{\text{out}}} = 0.25, \\
    \text{Output Weight} &= \frac{P_{\text{out}}}{P_{\text{in}} + P_{\text{out}}} = 0.75.
\end{align*}

Substituting these weights, the formula simplifies to:
\begin{equation}
    N_{\text{total}} = 0.25 \cdot N_{\text{in}} + 0.75 \cdot N_{\text{out}}.
\end{equation}

\subsection{Dataset-Specific Computational Cost Analysis}

The computational cost of Table-Critic is compared against Chain-of-Table for the WikiTQ and TabFact datasets. Detailed token counts and cost ratios are shown in Table~\ref{cost_comparison}.


On the WikiTQ dataset, Chain-of-Table incurs a total computational cost of $19.6$M, with $73.5$M input tokens and $1.6$M output tokens. In contrast, Table-Critic requires $135.5$M input tokens and $3.8$M output tokens, resulting in a total cost of $36.7$M. This corresponds to a cost ratio of $1.87\times$, indicating that Table-Critic is approximately $1.87$ times more computationally expensive than Chain-of-Table on this dataset.



On the TabFact dataset, Chain-of-Table incurs a total computational cost of $7.8$M, with $29.3$M input tokens and $0.6$M output tokens. Table-Critic, on the other hand, requires $62.1$M input tokens and $20.4$M output tokens, resulting in a total cost of $17.1$M. This corresponds to a cost ratio of $2.19\times$, indicating that Table-Critic is approximately $2.19$ times more computationally expensive than Chain-of-Table.

\section{Self-evolving Template Tree}\label{app:tree}
\begin{figure*}[t]
  \centering
  \includegraphics[width=\textwidth]{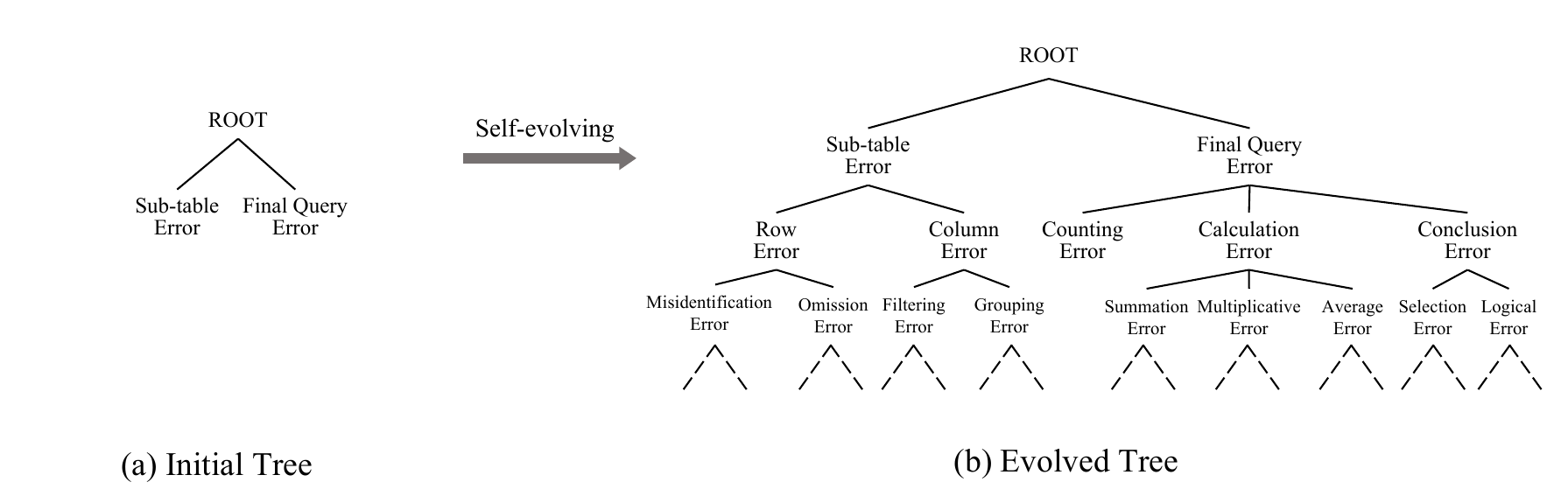}
  \caption{An example of self-evolving mechanism in our Template Tree.}
  \label{fig:tree}
\end{figure*}

Figure~\ref{fig:tree} illustrates the Self-evolving process of the Template Tree. In the initial stage (Figure~\ref{fig:tree}a), the tree contains only two broad categories of errors: Sub-table Error and Final Query Error, each representing a high-level abstraction of error types. Through the self-evolving mechanism, the tree dynamically expands and refines its structure to accommodate more fine-grained error types, as shown in the evolved tree (Figure~\ref{fig:tree}b).

It is important to note that the Evolved Tree is considerably larger in practice, containing a more extensive hierarchy of error types. However, for clarity, only a subset of the evolved structure is displayed here. 

\section{Prompts and Case Study}
\label{app:prompts}

This appendix provides comprehensive instructions and illustrative examples for four intelligent agents: the Judge Agent, the Critic Agent, the Refiner Agent, and the Curator Agent. These agents are designed to collaboratively evaluate and refine reasoning processes applied to table-based questions.

Figures~\ref{fig:judgmentinstructions} and~\ref{fig:judgmentexample} offer detailed guidance for the \textbf{Judge Agent}, including step-by-step procedures to assess the validity of reasoning steps, pinpoint errors, and categorize conclusions (e.g., correct, incorrect with identified error route, or random error). Figures~\ref{fig:critiqueinstructions} and~\ref{fig:critiqueexample} explain how the \textbf{Critic Agent} systematically evaluates each reasoning step, highlights the first incorrect step, and provides constructive critiques. Figure~\ref{fig:refinementexample} introduces the \textbf{Refiner Agent}, demonstrating how critiques are utilized to refine reasoning steps, ensuring accurate and complete solutions.

As for \textbf{Curator Agent}, Curator expands the template tree $\mathcal{T}$ by adding templates to existing leaf nodes or expanding tree branches (as described in Algorithm~\ref{alg2}). This process begins by invoking the \texttt{Judge} agent to determine the routing path and evaluation status of the refinement history within the context of the current template tree (Line 1). The outcome, denoted as $route$, indicates whether the refinement aligns successfully with any existing route in $\mathcal{T}$.

When the $route$ is successful within $\mathcal{T}$ ($route$.status = SUCCESS), Curator will perform Adding Templates or Vertical Expansion.
To make this decision, Curator uses a prompt to assess the similarity between the refinement history $H$ and the sampled templates $\mathcal{T}_r$ (see Figure~\ref{fig:curatorInstruction}). The agent then checks whether the refinement history $H$ is sufficiently similar to the node $T_r$ (Line 4).
If so (Figure~\ref{fig:addtemplate}), the template tree is directly augmented by adding the new template derived from $H$ through the \texttt{AddTemplate} function (Line 5).
If the refinement history diverges in content but follows a related path (Figure~\ref{fig:verticalexpansion}), a \texttt{VerticalExpansion} is performed to hierarchically extend the tree structure while preserving the routing path (Line 7).

Conversely, if the refinement path does not correspond to any existing route ($route$.status $\neq$ SUCCESS), the agent invokes \texttt{HorizontalExpansion} to create a new sibling branch in the template tree, thereby broadening its representational capacity (Line 10).
This horizontal expansion process is shown in Figure~\ref{fig:horizontalexpansion}, where a new branch is added to accommodate unmatched templates.

This process ensures that the template tree $\mathcal{T}$ evolves into a more expressive and comprehensive structure $\mathcal{T}'$, capable of supporting a wider range of reasoning refinements. The final evolved tree $\mathcal{T}'$ is returned as the output (Line 12), encapsulating the incremental knowledge acquired through iterative refinement.

\begin{algorithm}
\caption{The implementation of Curator Agent}
\renewcommand{\algorithmicrequire}{\textbf{Input:}}
\renewcommand{\algorithmicensure}{\textbf{Output:}}
\label{alg2}
\begin{algorithmic}[1]
\REQUIRE The template tree $\mathcal{T}$, the refinement history $H$.
\ENSURE The evolved tree $\mathcal{T}'$.

\STATE $route \leftarrow \texttt{Judge}(\mathcal{T}, H)$

\IF{$route.\text{status} = \text{SUCCESS}$}
    \STATE $\mathcal{T}_r \leftarrow$ Sample Templates using $route$ in the $\mathcal{T}$
    \IF{$H$ is similar to $\mathcal{T}_r$}
    \STATE $\mathcal{T}' \leftarrow \texttt{AddTemplate}(\mathcal{T}, H)$
    \ELSE
    \STATE $\mathcal{T}' \leftarrow \texttt{VerticalExpansion}(\mathcal{T}, H)$
    \ENDIF
\ELSE
    \STATE $\mathcal{T}' \leftarrow \texttt{HorizontalExpansion}(\mathcal{T}, H)$
\ENDIF

\RETURN Evolved template tree $\mathcal{T}'$
\end{algorithmic}
\end{algorithm}





\begin{figure*}[h]
\centering
\small
\begin{tcolorbox}[colframe=black, colback=white, coltitle=black, arc=4mm]
\parbox{15cm}{\ttfamily
You are an intelligent judge tasked with evaluating the correctness of a given Prediction Answer. If the Prediction Answer is incorrect, identify which step within the reasoning process is incorrect and subsequently locate the corresponding error type within the error tree: \par
1. Original Table: The raw table data. \par
2. Question: The question pertaining to the table data. \par
3. Reasoning Steps: A step-by-step process of sub-table transformations and extractions based on the following functions. \par
\ \ - f\_add\_column(): Adds a new column to the table. \par
\ \ - f\_select\_row(): Selects specific rows based on the question. \par
\ \ - f\_select\_column(): Removes irrelevant columns from the table. \par
\ \ - f\_group\_column(): Groups rows based on the values in a specific column. \par
\ \ - f\_sort\_column(): Sorts rows based on the values in a specified column. \par
4. Prediction Answer: The answer derived from the final sub-table. \par \par
\ \par
\textbf{Instruction:} \par
1. \textbf{Explanation:} Conduct an explanation of why the Prediction Answer is correct or incorrect. If it is incorrect, then conduct an analysis of each reasoning step's validity. \par
2. \textbf{Conclusion:} \par
\ \ - If the Prediction Answer is correct, conclude with `Conclusion: [Correct]'. \par
\ \ - If the Prediction Answer is incorrect, conclude with either `Conclusion: [Incorrect] (ERROR ROUTE)' or `Conclusion: [Incorrect] (random)'. \par
\ \ - Use `(ERROR ROUTE)' to indicate the specific path in the error tree that represents the error. \par
\ \ - If no such route can be identified, use `(random)' instead. \par
}
\end{tcolorbox}
\caption{Instructions for the Judge Agent. These instructions outline the procedure for verifying the correctness of a predicted answer and identifying errors within the reasoning process.}
\label{fig:judgmentinstructions}
\end{figure*}

\begin{figure*}[h]
\centering
\small
\begin{tcolorbox}[colframe=black, colback=white, coltitle=black, arc=4mm]
\parbox{15cm}{\ttfamily
Original Table: \par
/* \par
col   : res. | record | opponent         | method                        | event                                        | date               | round | time | location                                | notes \par
row 1 : win  | 12-3   | mike hayes       | ko (punch)                    | ksw 25: khalidov vs. sakurai                 | december 7, 2013   | 1     | 1:12 | wrocław, poland \par
row 2 : win  | 11–3   | nick moghadden   | tko (punches)                 | bellator 99                                  | september 13, 2013 | 1     | 3:22 | temecula, california, united states     | bellator debut \par
row 3 : loss | 10–3   | guto inocente    | decision (unanimous)          | strikeforce: barnett vs. cormier             | may 19, 2012       | 3     | 5:00 | san jose, california, united states     | light heavyweight debut \par
row 4 : win  | 10–2   | brett albee      | tko (strikes)                 | strikeforce: diaz vs. daley                  | april 9, 2011      | 1     | 1:46 | san diego, california, united states \par
row 5 : loss | 9–2    | lavar johnson    | ko (punches)                  | strikeforce challengers: bowling vs. voelker | october 22, 2010   | 1     | 2:17 | fresno, california, united states \par
row 6 : win  | 9–1    | eddie sapp       | submission (rear-naked choke) | native fighting championship 6               | august 14, 2010    | 1     | 2:01 | campo, california, united states \par
row 7 : loss | 8–1    | cody goodale     | decision (unanimous)          | gladiator challenge: maximum force           | april 25, 2010     | 3     | 5:00 | san jacinto, california, united states \par
*/ \par \par
Question: \par
how many loses did virgil zwicker have? \par \par
Reasoning Steps: \par
Step 1: Select relevant rows. \par
The question aims to find out the number of losses Virgil Zwicker had based on the provided fight records. We need to count the rows where the "res." column shows "loss". \par
So we use f\_select\_row(row 3, row 5, row 7). \par \par
Step 2: Filter out useless columns. \par
similar words of the question link to columns: \par
loses -> record \par
column value of the question link to columns: \par
10–3 -> record \par
semantic sentence of the question link to columns: \par
None \par
So we use f\_select\_column(record). \par \par
Step 3: After using f\_select\_row(row 3, row 5, row 7) and f\_select\_column(record), we obtain the sub table: \par
/* \par
col   : record \par
row 1 : 10–3 \par
row 2 : 9–2 \par
row 3 : 8–1 \par
*/ \par
The provided table shows records of wins and losses but does not specify who the records belong to. However, if we assume that the table represents Virgil Zwicker's records, then we can deduce the number of losses from each record. The losses in the records are 3, 2, and 1 respectively. If we consider each row as a separate season or period, then the total number of losses would be the sum of these losses: 3 + 2 + 1 = 6. \par \par
Prediction Answer: \par
6 \par \par
\ \par
\textbf{Explanation:} \par
Step 1 correctly identifies the relevant rows where the "res." column indicates a loss. The rows selected are indeed the ones where the outcome is a loss. Step 1 is correct. \par
Step 2 incorrectly filters out the columns. The question asks for the number of losses, but the filtering step selects only the 'record' column, which combines wins and losses in a single string (e.g., "10–3"). This does not directly provide the number of losses. Instead, the 'res.' column should be used to count the losses directly. Step 2 is incorrect. \par \par
\ \par
\textbf{Conclusion:} [Incorrect] (sub-table error -> column error -> <END>)
}
\end{tcolorbox}
\caption{Example of Judge Agent's Analysis and Error Detection. This example illustrates how the Judge Agent evaluates reasoning steps, identifies errors, and determines the correctness of a predicted answer.}
\label{fig:judgmentexample}
\end{figure*}

\begin{figure*}[h]
\centering
\small
\begin{tcolorbox}[colframe=black, colback=white, coltitle=black, arc=4mm]
\parbox{15cm}{\ttfamily
You are an intelligent critic tasked with determining which step of the table reasoning is incorrect based on the following information: \par
1. Original Table: The raw table data. \par
2. Question: The question pertaining to the table data. \par
3. Reasoning Steps: A step-by-step process of sub-table transformations and extractions based on the following functions. \par
\ \ - f\_add\_column(): Adds a new column to the table. \par
\ \ - f\_select\_row(): Selects specific rows based on the question. \par
\ \ - f\_select\_column(): Removes irrelevant columns from the table. \par
\ \ - f\_group\_column(): Groups rows based on the values in a specific column. \par
\ \ - f\_sort\_column(): Sorts rows based on the values in a specified column. \par
4. Prediction Answer: Final derived answer following the reasoning chain. \par \par
\ \par
\textbf{Instruction:} \par
1. \textbf{Step-wise Analysis}: Conduct an evaluation of each reasoning step's validity. The step that is unnecessary but does not affect the answer is considered correct. \par
2. \textbf{Analysis Categories}: \par
\ \ - For correct steps: Provide validation reasoning and mark as Step <NUM> is correct. \par
\ \ - For incorrect steps: Detail the logical flaws and mark as Step <NUM> is incorrect. \par
\ \ - You should stop at the first incorrect step. \par
3. \textbf{Conclude this critique}: Summarize this critique with an explicit conclusion. \par
4. \textbf{Conclusion Categories}: \par
\ \ - Conclude with 'Conclusion: [Incorrect] Step <NUM>'. \par
}
\end{tcolorbox}
\caption{Instructions for the Critic Agent. These instructions guide stepwise evaluation of reasoning steps.}
\label{fig:critiqueinstructions}
\end{figure*}

\begin{figure*}[h]
\centering
\small
\begin{tcolorbox}[colframe=black, colback=white, coltitle=black, arc=4mm]
\parbox{15cm}{\ttfamily
Original Table: \par
/* \par
col   : tie | home team              | score | away team          | attendance \par
row 1 : 1   | aylesbury united       | 2-2   | windsor \& eton     | 847 \par
row 2 : 2   | burscough              | 5-0   | wakefield \& emley  | 437 \par
row 3 : 3   | dover athletic         | 0-3   | forest green roves | 932 \par
row 4 : 4   | farnborough town       | 2-0   | halifax town       | 863 \par
row 5 : 5   | gloucester city        | 1-1   | southport          | 1,237 \par
row 6 : 6   | havant \& waterlooville | 3-0   | hayes              | 456 \par
row 7 : 7   | margate                | 0-2   | tamworth           | 971 \par
row 8 : 8   | yeovil town            | 2-1   | northwich victoria | 4,469 \par
*/ \par \par
Question: \par
how many games attendance was under 1000? \par \par
Reasoning Steps: \par
Step 1: Select relevant rows. \par
The question aims to find out how many games had an attendance of less than 1000 people. To answer this, we need to count the number of rows where the attendance column shows a number less than 1000. So we use f\_select\_row(row 1, row 2, row 3, row 4, row 6, row 7). \par \par
Step 2: Filter out useless columns. \par
similar words of the question link to columns : \par
under 1000 -> attendance \par
column value of the question link to columns : \par
1000 -> attendance \par
semantic sentence of the question link to columns : \par
None \par
So we use f\_select\_column(attendance). \par \par
Step 3: After using f\_select\_row(row 1, row 2, row 3, row 4, row 6, row 7) and f\_select\_column(attendance), we obtain the sub table: \par
/* \par
col   : attendance \par
row 1 : 847 \par
row 2 : 437 \par
row 3 : 932 \par
row 4 : 863 \par
row 5 : 456 \par
row 6 : 971 \par
*/ \par
The games with attendance under 1000 were rows 1, 2, 4, and 5. Therefore, there were 4 games with attendance under 1000. \par \par
Prediction Answer: \par
4 \par \par
\ \par
\textbf{Critique:} \par
Step 1 correctly selects the rows that are relevant for further analysis regarding the attendance being under 1000. It includes all the rows where the attendance values might potentially be less than 1000, which are row 1, row 2, row 3, row 4, row 6, and row 7. This step makes the proper initial identification based on the condition stated in the question. Step 1 is correct. \par
Step 2 filters out the columns effectively by choosing the "attendance" column. Since the question is focused on the attendance values to determine which are under 1000, this step narrows down the data to just what's needed for the subsequent steps. Step 2 is correct. \par
Step 3 makes an error in counting the number of games with attendance under 1000. Looking at the sub-table from Step 3, all the rows shown (row 1 with 847, row 2 with 437, row 3 with 932, row 4 with 863, row 5 with 456, row 6 with 971) have attendance values that are less than 1000. Step 3 is incorrect. \par 
\ \par
\textbf{Conclusion:} [Incorrect] Step 3 \par
}
\end{tcolorbox}
\caption{Example of Critic Agent's Critique. This example demonstrates how to evaluate reasoning steps and identify errors in the reasoning chain.}
\label{fig:critiqueexample}
\end{figure*}

\begin{figure*}[h]
\centering
\small
\begin{tcolorbox}[colframe=black, colback=white, coltitle=black, arc=4mm]
\parbox{15cm}{\ttfamily
Now, we have produced part of the Function Chain, but gained a critique. \par
Function Chain: f\_select\_row(row 1) \par
After step 1 (f\_select\_row(row 1)), we obtain the sub-table: \par
/* \par
\ \ col : date introduced | class 1 (e.g. motorbike) | class 2 (e.g. car) | class 3 (e.g. car with trailer) | class 4 (e.g. van) | class 5 (e.g. hgv) \par
\ \ row 1 : 23 july 2004 | £1.00 | £2.00 | £5.00 | £5.00 | £6.00 \par
*/ \par
Question: on what date did the toll for class 1 first go above 2.00? \par
Critique: \par
Step 1 is incorrect. The selected row (row 2) has the toll for class 1 set at £1.00, which is not above £2.00. The first row where the toll for class 1 exceeds £2.00 is row 3, dated 16 august 2004. Therefore, the selection of row 2 is incorrect. \par
\ \par
Based on the \textbf{critique}, please continue to produce a complete and correct Function Chain. \par
/* \par
\ \ col : date introduced | class 1 (e.g. motorbike) | class 2 (e.g. car) | class 3 (e.g. car with trailer) | class 4 (e.g. van) | class 5 (e.g. hgv) \par
\ \ row 1 : 9 december 2003 | £1.00 | £2.00 | £5.00 | £5.00 | £10.00 \par
\ \ row 2 : 23 july 2004 | £1.00 | £2.00 | £5.00 | £5.00 | £6.00 \par
\ \ row 3 : 16 august 2004 | £2.00 | £3.00 | £6.00 | £6.00 | £6.00 \par
\ \ row 4 : 14 june 2005 | £2.50 | £3.50 | £7.00 | £7.00 | £7.00 \par
\ \ row 5 : 1 january 2008 | £2.50 | £4.50 | £8.00 | £9.00 | £9.00 \par
\ \ row 6 : 1 january 2009 | £2.70 | £4.70 | £8.40 | £9.40 | £9.40 \par
\ \ row 7 : 1 march 2010 | £2.70 | £5.00 | £9.00 | £10.00 | £10.00 \par
\ \ row 8 : 1 march 2011 | £3.00 | £5.30 | £9.60 | £10.60 | £10.60 \par
\ \ row 9 : 1 march 2012 | £3.00 | £5.50 | £10.00 | £11.00 | £11.00 \par
*/ \par
Question: on what date did the toll for class 1 first go above 2.00? \par
The next operation must be one of f\_add\_column(), f\_select\_row(), f\_select\_column(), f\_group\_column(), or f\_sort\_column(). \par
\ \par
\textbf{Function Chain:} \par
f\_select\_row(row 3) \par
}
\end{tcolorbox}
\caption{Example of Refiner Agent's refinement. This example demonstrates how the critique is used to refine the Function Chain to accurately answer the question.}
\label{fig:refinementexample}
\end{figure*}

\begin{figure*}[h]
\centering
\small
\begin{tcolorbox}[colframe=black, colback=white, coltitle=black, arc=4mm]
\parbox{15cm}{\ttfamily
You are organizing hierarchical categories and their associated few-shot examples. Currently, you have two lists of few-shot examples under the same category. Your task is to decide whether these two lists can be meaningfully split into two distinct subcategories. \par

\textbf{Instructions:} \par
Analyze the examples in the two lists to determine if there is a clear and meaningful distinction between them. \par
\ \ - If a distinction exists, create two subcategories and assign each list to one of them. \par
\ \ - If no clear distinction exists, retain both lists under the original parent category. \par
Provide a clear explanation for your decision on whether to split or merge the lists, based on their content. \par

\textbf{Parent Category:} row error \par

\textbf{List 1:} \par
[ \par
\ \ (Template 1), \par
\ \ (Template 2), \par
\ \ ... \par
] \par

\textbf{List 2:} \par
[ \par
\ \ (New Template) \par
] \par

}
\end{tcolorbox}
\caption{Instructions for determining whether the refinement history $H$ is similar to sampled templates $\mathcal{T}_r$. It guides the model to decide if the new templates should be merged with the existing category or split into distinct subcategories.}
\label{fig:curatorInstruction}
\end{figure*}

\begin{figure*}[h]
\centering
\small
\begin{tcolorbox}[colframe=black, colback=white, coltitle=black, arc=4mm]
\parbox{15cm}{\ttfamily
\textbf{Explanation:} \par
... \par

\textbf{Determination:} \par
List 1: <row error> \par
List 2: <row error> \par
}
\end{tcolorbox}
\caption{Model explanation and categorization result when determining whether to add the refinement history $H$ to existing templates $\mathcal{T}_r$. This corresponds to the \texttt{AddTemplate} operation in Algorithm~\ref{alg2}.}
\label{fig:addtemplate}
\end{figure*}

\begin{figure*}[h]
\centering
\small
\begin{tcolorbox}[colframe=black, colback=white, coltitle=black, arc=4mm]
\parbox{15cm}{\ttfamily
\textbf{Explanation:} \par
... \par

\textbf{Determination:} \par
List 1: <row misidentification error> \par
List 2: <row omission error> \par
}
\end{tcolorbox}
\caption{Model explanation and categorization decision when refinement history $H$ is not exactly similar to existing templates $\mathcal{T}_r$. This corresponds to the \texttt{VerticalExpansion} step in Algorithm~\ref{alg2}.}
\label{fig:verticalexpansion}
\end{figure*}

\begin{figure*}[h]
\centering
\small
\begin{tcolorbox}[colframe=black, colback=white, coltitle=black, arc=4mm]
\parbox{15cm}{\ttfamily
You are given an error tree represented as a dictionary and a template describing a specific error. \par
If the error path corresponding to the template cannot be found in the error tree, extend the error tree by adding a new branch at the appropriate location. \par
Ensure that: \par
\ \ - The new branch aligns logically with the existing structure of the error tree. \par
\ \ - The template can be correctly integrated into the tree under the newly added branch. \par

\textbf{Error Tree:} \par
\{ \par
\ \ \ "sub-table error": \{ \par
\ \ \ \ \ \ "row error": "<END>", \par
\ \ \ \ \ \ "column error": "<END>" \par
\ \ \ \} \par
\} \par

\textbf{Template:} \par
(New Template) \par
------------------------------------------------------------------------------------------ \par
For example, if the return result is as follows: \par
\textbf{Addition: (final query error -> <END>)} \par

Curator will add a "final query error" node under the root node for this new template. \par

}
\end{tcolorbox}
\caption{Instructions and example of \texttt{HorizontalExpansion}. Prompt guiding the model to extend the template tree horizontally by adding a new branch when the refinement history $H$ does not match any existing template paths. This corresponds to the \texttt{HorizontalExpansion} step in Algorithm~\ref{alg2}.}
\label{fig:horizontalexpansion}
\end{figure*}

\end{document}